# ONLINE FAKE REVIEW DETECTION USING SUPERVISED MACHINE LEARNING AND BERT MODEL

[1]Abrar Qadir Mir, [2]Furqan Yaqub Khan, Mohammad Ahsan Chishti

[1,3]Dept. of IT, CU Kashmir,

J&K, India, 191131

[2]Dept. of CSE, IIT Patna, Bihar, India 801106

*Corresponding Author: furkaan309@gmail.com*

## Abstract

Online shopping stores have grown steadily over the past few years. Due to the massive growth of these businesses, the detection of fake reviews has attracted attention. Fake reviews are seriously trying to mislead customers and thereby undermine the honesty and authenticity of online shopping environments. So far, various of fake review classifiers have been proposed that take into account the actual content of the review. To improve the accuracies of existing fake review classification or detection approaches, we propose to use BERT (Bidirectional Encoder Representation from Transformers) model to extract word embeddings from texts (i.e. reviews). Word embeddings are obtained in various basic methods such as SVM (Support vector machine), Random Forests, Naive Bayes and others. The confusion matrix method was also taken into account to evaluate and graphically represent the results. The results indicate that the SVM classifiers outperforms the others in terms of accuracy and f1-score with an accuracy of 87.81%, which is 7.6% higher than the classifier used in the previous study [5].

## 1. Introduction

Online reviewing play vital role in various online shopping environments such as online hotel reservations, restaurant reservations and more importantly in the case of e-commerce sites as they make it possible for customers to judge the quality/price of different products in order. evaluate the seller's services and also make purchasing decisions. The users using online shopping platforms has been increasing significantly for quite some time now. Let's take an example of TripAdvisor, globally the most popular travel website, has approximately 455.5 million users and more than 600 million user reviews covering more than 7.5 million hotel, restaurant, and airline reservations [1]. However, some unscrupulous businesses, motivated by self-interest, mislead some less cautions customers by writing fake reviews. In order to over promote products of their choice while disparaging their alternate choices. Fake reviewing drastically affect and threaten the authenticity of online shopping environments. Therefore, detecting and removing such fake reviews has become vital and unconditionally important area of research for the present and if not dealt properly possibly for future as well. The Fake review detection can be manual or automatic but manual execution of the process is considered expensive, time-consuming and incorrect compared to automated detection methods [2]. Over the past decade, many significant advances have been made in the automatic identification of fake reviews. Many ML based tools and techniques have been effectively detecting fake reviews, some of the tools and techniques are SVM and Naïve Bayes[3]. This study mainly focuses on review content to detect fake reviews. NLP is used to create some review features which are not directly related to data or text provided. We used BERT (Bidirectional Encoder Representation from Transformers) to perform NLP tasks and other text processing such as feature selection.

## 2. Related work

The online shopping environment has started to gain massive interest in recent past and plays a important role in people's lives, saving time or saving money. People seem to rely quite heavily on such online shopping. From this point of view, finding out online reviews is equally important and is one of the hot research topics today. The problem of detecting fake reviewing had been addressed since 2007 [4]. The basic categories for fake review detection research are categorized as; textual and behavioural features. Text characters means characterization of text of reviews or to analyse and characterise the content of review text. Behavioural traits characterise the non-verbal characteristics of review data corpus. It mainly depends on the behaviour of reviewers, like style of writing, expressions which express emotions, and frequency of reviewing. Liu [6] proposed the first review detection model using review similarity and products features as criteria. The quality of fake reviews, also called "spammers", to post fake reviews of their products has been exploited. Ott [7] proposed a classification model based on text polarity of review data corpus; created a dataset that contained four-hundred reviews divided into 2 groups real and fake or artificial and calculated the effectiveness of his model using different classifiers and on different system configurations. Savage [8] improved the efficiency of Ott's model by updating it with some new syntactic features. Farris [9] proposed another model for classification of fake reviews using Genetic Algorithm with Random Weight Network. This model specifically addressed spam detection. This model generated excellent results and it was concluded with this model that spam reviews can be classified with good precision, recall and accuracy using automated classifiers. Fake reviewer identification: [10] Although the above methods worked well in detecting fake comments and reviews, it was recognized that fake reviewer detection is an additional part and needs to be invented. Huang [11] came up with another theory that inauthentic reviewers usually write comments frequently in particular time periods, so he defined time frame for comment upload time and estimated upload frequencies of fake comments which where unusually different from real reviews, i.e.; **If,** uploads > threshold; The reviewer is said to be fake.

Lim [12] examined a large number of reviewers and similar data and pointed out two features of fake reviewers: 1) fake reviewers target review a "particular" product or a specific online purchase. 2) The pattern and style of writing and the frequency of posting reviews is generally different for a real and a fake reviewer. Based on these two complimentary traits of reviewing, it modelled the reviewer's behaviour and calculated the reviewer's score using another algorithm. In [13], Supervised ML were used to classify reveiws. The supervised classification techniques used are Naïve-Bayes, SVM, K-Star, KNN, and Decision trees. The experimentation was performed on three different movie review labelled datasets.[14] having 1400, 2000, 10662 reviews per dataset. In [15], the Naïve-Bayes, SVM, Decision Tree techniques were used along with classifiers like Random Forest and Maximum Entropy Classifier. The dataset forming of Samsung products and services reviews with 10000 fake reviews. In [16], Chicago hotel review dataset of 1600 reviews was used. In [17], The RNN, Average GRNN, GRNN, CNN, and Bi-directional Average GRNN's were used to classify data corpus. The dataset used was from [18] having both fake and real reviews based on hotels, restaurants and doctors. Only textual features were taken into consideration discarding behavioural features.

## 3. Background

To parse the high volume data with velocity which can't be dealt through routine algorithms for time critical tasks with accuracy machine learning is tool of choice. Machine Learning has an advantage of learning and developing algorithms on its own based on data patterns and improvise them while in use with further data[19]. The classification of machine learning algorithms is based on labels and action-reward theory[20]; Based on Labels the classes are supervised, semi-supervised and unsupervised ML techniques while based on action-reward theory reinforced machine learning techniques are there. Labels and data are both needed in supervised machine learning techniques[21] while in unsupervised ML only data is provided and relation is found between different data points based on desired attributes or functions. The semi-supervised techniques are hybrid of supervised and unsupervised ML techniques. Finally, Reinforced learning deals with reacting to particular scenario or action if the reaction is in the right direction the priority is added to step taken that is the step is rewarded if not the priority is decreased for that particular step that means the step is punished. Mainly Supervised ML classifiers are used in this article. Few prominent supervised classifiers used are SVM which separates the two needed classes via a hyperplane[22] Another one is naïve bayes classifier which uses bayes theorm to get the probability of particular review to be of one of the given classes. The equation for bayes theorm is $P(A—B) = P(B—A)*P(A) P(B)$ [23].

The K-Nearest Neighbour's (KNN) algorithm [24] mostly used in statistical estimations and as well in pattern recognitions. The distance function is used to classify the least distant attributes into one cluster thus trying to minimize the intra-cluster distance and maximise the inter-cluster distances in order to classify more accurately[25] Decision Tree [26] classifies data based values of gini-ratio, entropy, information gain etc. Thus each decision point is represented as an internal node of tree while each class is represented by leaf node of tree giving the algorithm's representation a tree like shape hence called decision tree. Random Forest [27] is an ensemble of more than one decision trees usually all unique thus eliminating the problem of overfitting which usually occurs in decision tree algorithm. Logistic regression [28] also develops a hyperplane between two different datatypes based on logistic function or log function.

## 4. Data files

Primarily four datasets viz hotel, doctor, restaurant and amazon datasets are used in this article for experimentation(Table 1). Cornell Universities Positive review[29] and negative review[30] datasets are merged to form hotel dataset as described in[29-30].These datasets are considered the gold standard mock review data corpuses[30]. Turks or assigned Fake reviewers were the ones responsible for creating fake reviews, one review per Turk. The reviews were filtered as short and plagiarized review were removed. Turkers were instructed to write a review which seems realistic in nature. Correspondingly geniune reviews were taken from various online reviewing sites like Expedia, TripAdvisor and Hotels.com. The dataset contains reviews of 20 hotels with 80 reviews of each one divided equally into 2 classes thus forming 1600 reviews half of class true reviews and half of class false reviews. Corresponding actual true and false labels were also included for each review. Thus each review in the dataset has a true/false label, hotel information, travel agency name, polarity (positive/negative), and review content with average word count of 152 per review. Similarly two more datasets were created [31] namely restaurant review dataset and doctor review dataset. Twenty fake reviews were created for 10 most famous restaurants in Chicago and 356 positive fake reviews for doctor review dataset were created by well rated American Turkers and true reviews of same number were taken from customer reviews of those hotels and doctors. Amazon's dataset consists of 21,000 reviews, of which 10,500 were identified as fake by Amazon. Some additional information was also provided about the reviews like class label, ratings, verified purchase (yes or no), product category, and product ID. With average review rating as 4.13/5 as well with 55.7% of the data from verified purchases. The reviews were form 30 product categories 700 reviews per category. These categories are identified as non-compliant with Amazon's policies.

Table 1 Fake review datasets used in this study

| Dataset | Fake/truthful reviews | Polarity | Aver. review length (words) |
| --- | --- | --- | --- |
| Hotel [29, 30] | 400/400 | Positive and negative | 151.9 |
| Restaurant [31] | 200/200 | Positive | 137.1 |
| Doctor[32] | 356/200 | Positive | 102.4 |

## 5. The proposed system

The proposed approach shown in Figure 1. The approach is divided into 3 stages and provides us with the best model for fake review classification given as follows:

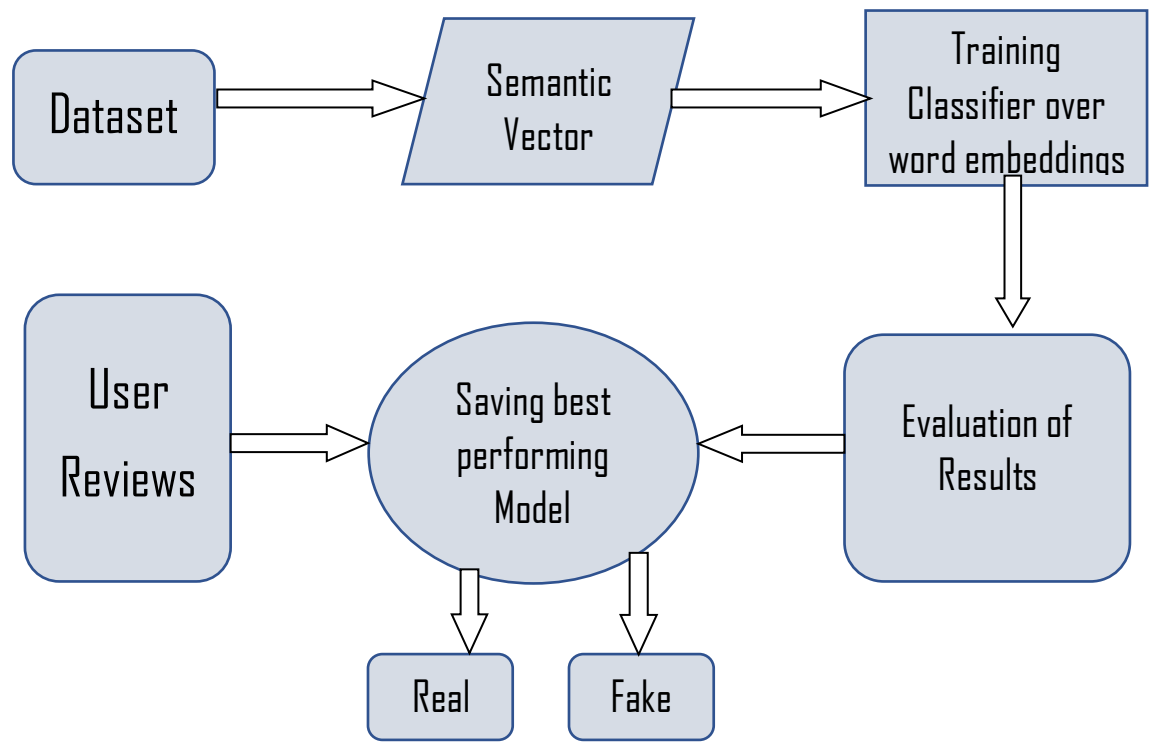

Fig 1: Working of the proposed model

1) The dataset is loaded into a BERT (Bidirectional Encoder Representation from Transformers) to generate word embeddings, which are large vector representations of the text words in the dataset. 2) The input is then loaded into the classification models for their training. Training and testing data are divided in the ratio: 80:20, i.e.: 80% for training and the rest for testing the model. 3) Results are evaluated using a confusion matrix representation for Precision, Recall, F1score and Accuracy. 4) The best performing classifier is then saved and later used to detect user reviews as fake or real.

## 6. Operation of the BERT model

BERT (Bidirectional Representations from Transformers) based on transformer – an attention mechanism that learns contextual connections in text between words. A simple transformation has an encoder to read the text input and a decoder to generate the task prediction. Since the goal of BERT is to create a model of the language representation, it only requires part of the coder. The input to BERT encoder is a set of tokens that are first transformed into vectors and then processed in a neural network. However, BERT requires the input to be decorated with some additional metadata before processing can begin. Transformer essentially composes a layer that maps onto sequences of sequences, so the output is also a vector sequence.

### 6.1 Generating BERT Embedding

Generating a large sentence embedding is created using BERT. BERT is a pre-trained model. Let the sentence $S_i$ be tokenized into words $W = \{w_1, w_2, w_3,..,w_n\}$. Each $w \in W$ is fed to

BERT to obtain the word embeddings for $w_s \in W$. Let the loaded embedding for w1 be E1. Similarly, we obtain embeddings for all wn words of Si, and so we have Vn embeddings for n words of sentence Si. Then all embeddings of the Si theorem are combined to form the large embeddings (BV) of the Si theorem. So BE = {E1 ∪ E1 ∪ E1 ∪ ... ∪ En }. Inserts are generated for the hotel dataset. Once we have all large embeddings of all reviews and their corresponding labels, we load these embeddings into the classification models for training and testing, in ratio of 80:20 training/testing data.

## 7. Experimental evaluation:

In this section, we present the results of six experiments and their evaluation using six different machine learning classifiers, namely: SVM, Random Forest, Bagging classifier, AdaBoost classifier, Naïve Bayes and K-NN classifier. We experimented with our classification model on the Hotel, Restaurant and Doctor datasets. However, due to space constraints, we only show the experimental results performed on the hotel dataset. The confusion matrices for all implemented classifiers are given below:

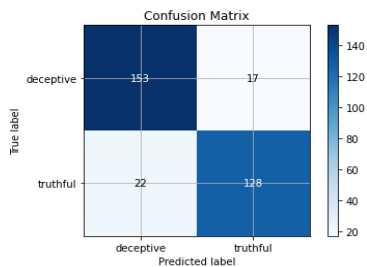

Figure 2: Confusion matrix for SVM

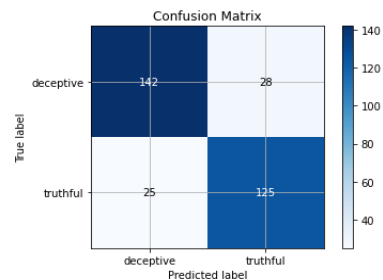

Figure 3: Confusion matrix for Random Forest

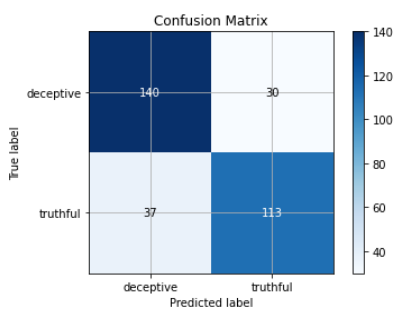

Figure 4: Confusion matrix for Bagging classifier

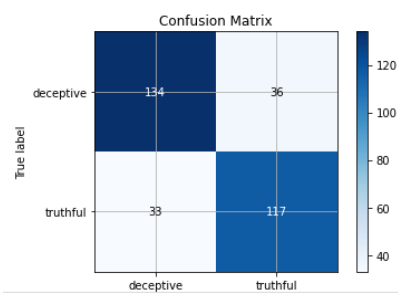

Figure 5: Confusion matrix for Adaptive Boost

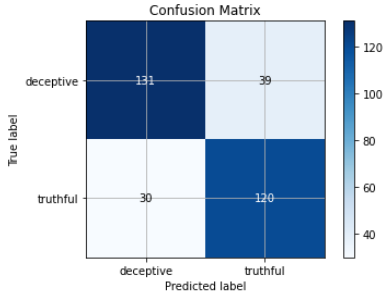
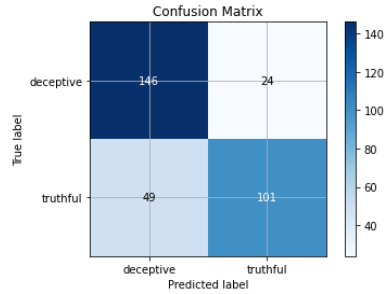

Figure 6: Confusion matrix for Naïve Bayes    Figure 7: Confusion matrix for K-NN classifier

| Classifier | Accuracy (%) | | F-Score | |
| --- | --- | --- | --- | --- |
| | Previous work[18] | This Study | Previous work[18] | This Study |
| **SVM** | 80.75 | **87.81** | 0.80 | **0.88** |
| **Random Forest** | 79.31 | **83.43** | 0.79 | **0.83** |
| **Bagging** | 78.19 | **79.06** | 0.78 | **0.79** |
| **K-NN** | 71.38 | **77.18** | 0.68 | **0.78** |
| **AdaBoost** | 77.06 | **78.43** | 0.77 | **0.78** |
| **Gaussian Naïve Bayes** | 81.25 | 78.43 | 0.81 | 0.78 |

Table 2: Comparative performance analysis of proposed model

## 7. Conclusion and future work

In this article, we've shown the importance of reviews and how they affect almost everything related to web data. It's clear that reviews play a vital role in people's decision-making. Thus, detecting fake reviews is a lively and ongoing area of research. A machine learning approach to detect fake reviews is presented, and review properties are considered in the proposed approach. The Hotels dataset is used to present an experimental evaluation of the proposed approach. Different classifiers were used on this dataset. The results reveal that the SVM classifier outperforms other classifiers (with 87.81% accuracy) in the process of detecting fake reviews and therefore can be used to effectively classify reviews as real or fake by considering only the text content of the reviews and not necessarily the sentiment traits. However, future work may consider including behavioural features of reviewers, such as features that depend on the number of times reviewers perform reviews, the time it takes reviewers to complete reviews, and how often they submit positive or negative reviews. Considering behavioural features is highly expected to enhance the performance of the presented fake review detection approach. Also, using neural network models to perform this task would be equally beneficial for detecting fake reviews from large datasets.

## References

1. TripAdvisor Homepage. http://ir.tripadvisor.com/. Accessed 21 January 2019.
2. Harris, C.: Detecting deceptive opinion spam using human computation. In: Workshops at AAAI on Artificial Intelligence, pp. 87–93. AAAI (2012)
3. Heydari, A., ali Tavakoli, M., Salim, N., Heydari, Z.: Detection of review spam: a survey. Expert Syst. Appl. 42(7), 3634–3642 (2015)
4. N. Jindal and B. Liu, "Review spam detection," in Proceedings of the 16th International Conference on World Wide Web, ser. WWW '07, 2007.
5. Hajek, P., Barushka, A. and Munk, M., 2020. Fake consumer review detection using deep neural networks integrating word embeddings and emotion mining. *Neural Computing and Applications*, *32*(23), pp.17259-17274.
6. Jindal N, Liu B (2007) Analyzing and detecting review spam. In: 7th IEEE international conference on data mining, ICDM 2007, IEEE, pp 547–552. https://doi.org/10.1109/icdm.2007.68
7. Ott M, Choi Y, Cardie C (2011) Finding deceptive opinion spam by any stretch of the imagination. Proc 49th Ann Meet Assoc Comput Linguist 6(8):309–319
8. Savage D, Zhang X, Chou P (2015) Detection of opinion spam based on anomalous rating deviation. Expert Syst Appl Int J 42(22):8650–8657
9. Faris H, Ala'm AZ, Heidari AA, Aljarah I, Mafarja M, Hassonah MA, Fujita H (2019) An intelligent system for spam detection and identification of the most relevant features based on evolutionary random weight networks. Inf Fusion 4:67–83
10. Fang, Y., Wang, H., Zhao, L., Yu, F. and Wang, C., 2020. Dynamic knowledge graph based fake-review detection. *Applied Intelligence*, *50*(12), pp.4281-4295.
11. Huang J, Qian T, He G (2013) Detecting professional spam reviewers, advanced data mining and applications. Springer, Berlin, pp 288 299
12. Lim EP, Nguyen VA, Jindal N, Liu B, Lauw HW (2010) Detecting product review spammers using rating behaviors. ACM Int Conf Inf Knowl Manag 7(12):939–948
13. E. Elmurngi and A. Gherbi, Detecting Fake Reviews through Sentiment Analysis Using Machine Learning Techniques. IARIA/DATA ANALYTICS, 2017.
14. V. Singh, R. Piryani, A. Uddin, and P. Waila, "Sentiment analysis of movie reviews and blog posts," in Advance Computing Conference (IACC), 2013, pp. 893–898.
15. A. Molla, Y. Biadgie, and K.-A. Sohn, "Detecting Negative Deceptive Opinion from Tweets." in International Conference on Mobile and Wireless Technology. Singapore: Springer, 2017.
16. S. Shojaee et al., "Detecting deceptive reviews using lexical and syntactic features." 2013.

17. Y. Ren and D. Ji, "Neural networks for deceptive opinion spam detection: An empirical study," Information Sciences, vol. 385, pp. 213–224, 2017.
18. H. Li et al., "Spotting fake reviews via collective positive-unlabeled learning." 2014.
19. D. Michie, D. J. Spiegelhalter, C. Taylor *et al.*, "Machine learning," *Neural and Statistical Classification*, vol. 13, 1994.
20. T. O. Ayodele, "Types of machine learning algorithms," in *New ad-vances in machine learning*. InTech, 2010.
21. F. Sebastiani, "Machine learning in automated text categorization," *ACM computing surveys (CSUR)*, vol. 34, no. 1, pp. 1–47, 2002.
22. T. Joachims, "Text categorization with support vector machines: Learn-ing with many relevant features." 1998.
23. T. R. Patil and S. S. Sherekar, "Performance analysis of naive bayes and j48 classification algorithm for data classification," pp. 256–261, 2013.
24. M.-L. Zhang and Z.-H. Zhou, "Ml-knn: A lazy learning approach to multi-label learning," *Pattern recognition*, vol. 40, no. 7, pp. 2038–2048, 2007.
25. N. Suguna and K. Thanushkodi, "An improved k-nearest neighbor clas-sification using genetic algorithm," *International Journal of Computer Science Issues*, vol. 7, no. 2, pp. 18–21, 2010.
26. M. A. Friedl and C. E. Brodley, "Decision tree classification of land cover from remotely sensed data," *Remote sensing of environment*, vol. 61, no. 3, pp. 399–409, 1997.
27. A. Liaw, M. Wiener *et al.*, "Classification and regression by random-forest," *R news*, vol. 2, no. 3, pp. 18–22, 2002.
28. D. G. Kleinbaum, K. Dietz, M. Gail, M. Klein, and M. Klein, *Logistic regression*. Springer, 2002.
29. Ott M, Cardie C, Hancock J (2012) Estimating the prevalence of deception in online review communities. In: 21st international conference on world wide web, ACM, pp 201–210. https://doi.org/10.1145/2187836.2187864
30. Ott M, Cardie C, Hancock JT (2013) Negative deceptive opinion spam. In: 2013 conference of the North American chapter of the association for computational linguistics: human language tech-nologies, ACL, pp 497–501
31. Li J, Ott M, Cardie C, Hovy E (2014) Towards a general rule for identifying deceptive opinion spam. In: Proceedings of the 52nd annual meeting of the association for computational linguistics, ACL, vol 1, pp 1566–1576.
32. Garcia L (2018) Deception on Amazon—an NLP exploration. https://medium.com/@lievgarcia/deception-on-amazon-c1e30d977cfd. Accessed 01 Sept 2019